# CreateScore: Domain-Theory-Informed Bayesian Routing for LLM-Based CV Screening

Rupsa Roy, MSc*

*Independent Researcher, MSc in Statistics (2019-21), National University of Singapore (NUS)

**Abstract**

Large language models (LLMs) can support rubric-based screening of curriculum vitae (CVs), but applying a high-capability model to every candidate and criterion imposes substantial computational and monetary costs. We present CreateScore, a domain-theory-informed Bayesian network for criterion-level LLM routing. CreateScore uses a hand-specified directed acyclic graph with Dirichlet–multinomial conditional probability tables to combine structured domain assumptions and observed CV evidence into posterior uncertainty estimates. The graph has a causally motivated structure, but the present system performs standard Bayesian conditioning rather than formal causal inference, intervention analysis, or counterfactual estimation. Criterion decisions with sufficiently low uncertainty are resolved by a low-cost 8B LLM, whereas uncertain decisions are escalated to a higher-capability 120B reference LLM. The escalation threshold is calibrated on a training subset to target 70% local resolution and then fixed for evaluation. We assess CreateScore using 200 synthetic Data Science CVs obtained from Kaggle and a five-criterion job order, with a stratified 70/30 candidate-level split yielding 139 training and 61 held-out test candidates. On the test set, 77.7% of criterion-level decisions were resolved locally (237 of 305), and 17 of 61 candidates required no reference-model call. Against a complete reference condition in which the 120B model adjudicated every criterion, routed escalation reduced token expenditure by 65.2% and raised exact score agreement from 32.8% (8B model alone) to 42.6% (95% CI 31.0–55.1%), with mean bias falling from +1.80 to +0.66 points on a 30-point scale; at n = 61 the gain in agreement was not statistically distinguishable. The posterior uncertainty signal did not, however, identify the decisions on which the 8B model erred: 8B–reference disagreement was 16.2% among escalated and 19.4% among locally resolved decisions (router AUROC 0.47, 95% CI 0.39–0.56), no better than selecting the same number of decisions at random. These findings support CreateScore, in its present form, as an auditable cost-reduction mechanism whose uncertainty signal requires improvement before it can claim to target ambiguity; it is not an autonomous hiring system. Validation with human-labelled, heterogeneous CV data and subgroup-specific fairness analyses remains necessary.

## 1 Introduction

### 1.1 Background

Large language models (LLMs) are increasingly being investigated for employment-related tasks, including information extraction, candidate summarization, and rubric-based evaluation of free-text curricula vitae (CVs). In a rubric-based screening workflow, an LLM may be required to assess several criteria for each candidate, such as professional experience, technical proficiency, and research background. Although a high-capability model may provide strong performance across these tasks, applying it uniformly to every candidate and criterion can produce substantial computational cost, monetary expense, and latency.

A cost-efficient alternative is a model cascade or routing architecture in which a lower-cost model handles routine cases and a stronger model is invoked only when the initial decision is uncertain. FrugalGPT introduced a cascade framework that selects among multiple LLMs or sequences of LLM calls to improve the quality–cost trade-off of inference (Chen et al., 2023). RouteLLM subsequently formulated model selection as a routing problem in which a learned router chooses between weaker and stronger LLMs using preference data and a cost–quality objective (Ong et al., 2024). These approaches demonstrate that selective model use can reduce inference expenditure while preserving task performance.

However, such methods share two structural assumptions: they operate on a complete query, and they treat escalation as binary, so that an input is either answered by the cheaper model or passed, in full, to the stronger one. A recent survey organizes routing and cascading methods by when the escalation decision is made rather than by what is transmitted once it has been made (Moslem & Kelleher, 2026), and a recent decision-theoretic treatment characterizes the frontier of whether to escalate without addressing the content of the escalated call (Bouchard, 2026). Two consequences follow. A decision taken for an entire query cannot represent the common case in which a candidate is unambiguous on some rubric criteria and genuinely uncertain on others. And if every escalation transmits the full input, the token cost of the stronger model remains largely decoupled from how selectively that model is invoked.

These limitations matter in employment screening, where an automated assessment may affect a person's access to work. Research on algorithmic hiring has shown that developers and vendors must address not only predictive performance but also validation, bias assessment, transparency, and scrutiny of how assessment systems are constructed and used (Raghavan et al., 2020). Recent empirical work further indicates that LLM-based resume screening and evaluation can produce race-, gender-, and intersection-specific differences even when candidates are otherwise comparably qualified (Wilson & Caliskan, 2024; An et al., 2025). A routing mechanism deployed in this setting should therefore be inspectable at the level of the individual decision it defers.

In this study, the high-capability model serves as a reference evaluator for the outputs of a lower-cost pipeline. This follows the LLM-as-a-judge paradigm, in which a strong LLM acts as a scalable proxy for human evaluation. Zheng et al. (2023) demonstrated that strong LLM judges can show substantial agreement with human judgments, while also documenting potential position, verbosity, and self-enhancement biases. Accordingly, agreement with the reference LLM in the present study should not be interpreted as ground-truth correctness, job-related validity, or evidence of fairness.

**1.2 Motivation**

The central motivation for CreateScore is that the routing signal in a cascade need not be learned in order to be useful. Rather than training a router end-to-end, CreateScore derives its escalation decisions from a small, hand-specified Bayesian network represented by a directed acyclic graph (DAG) (Pearl, 1988), with conditional probability tables parameterized using Dirichlet priors and fixed, hand-elicited pseudo-counts scaled by a global strength parameter. The machinery is deliberately conventional; the design claim is not that this class of model is new, but that placing a textbook probabilistic model at the routing layer supplies three properties a learned router does not. It can be specified from domain knowledge rather than estimated from labelled preference data, so that only a single scalar threshold is fitted. Its output is a posterior over a named criterion rather than an opaque score, so that the reason for deferral can be inspected. And because each criterion carries its own posterior, escalation can be decided independently for each criterion of a single candidate.

The graph is informed by domain assumptions concerning candidate seniority, skill–background relationships, and research-depth patterns, the last following cumulative-advantage accounts of scientific productivity (Merton, 1968). Expressed as priors rather than learned weights, these assumptions remain legible and can be revised or contested without retraining.

CreateScore should be understood as a domain-theory-informed Bayesian network with a causally motivated structure. The edge directions encode a proposed generative interpretation of how latent candidate characteristics may relate to observed CV evidence, but the present study performs standard Bayesian conditioning and posterior uncertainty estimation. It does not estimate causal effects, perform interventions using the do-operator, adjust for confounding, or generate counterfactual predictions (Pearl, 2009). The term "causally motivated" therefore refers to the interpretation and construction of the graph, not to a formal causal-inference procedure. The same distinction separates CreateScore from a hierarchical Bayesian model in the strict statistical sense: it does not estimate group-level parameters from a shared hyperprior, perform partial pooling, or learn hyperparameters across strata. Its layered structure is graphical rather than multilevel.

CreateScore also separates two decisions that cascade architectures ordinarily combine. The first is whether an individual criterion-level decision should be resolved locally or escalated, which it determines from posterior uncertainty. The second is how much and what type of CV context an escalated decision receives, treated here as an independent design axis rather than a property of the network. Holding the first fixed while varying the second would make it possible to attribute differences in agreement and token cost to the escalation payload alone; the corrected evaluation reported here tests a single payload policy, so that comparison is left to future work (Section 5.3).

The intended trajectory of this design extends beyond the present evaluation in two directions. Dirichlet posteriors concentrate as observed counts accumulate, so a larger corpus should yield sharper per-criterion posteriors and, at a fixed threshold, a higher proportion of decisions resolved locally; the savings available from routing are therefore expected to grow with corpus size rather than remain constant. Separately, a rubric substantially larger or more heterogeneous than the one used here is unlikely to be served well by a single hand-elicited structure. The intended extension is a set of smaller, connected DAGs specialized by job family or criterion group, which would localize uncertainty more precisely than one global graph and, in doing so, permit narrower and less expensive escalation payloads. Both are design expectations rather than findings, and neither is tested here.

This design is relevant to employment screening because interpretability must extend beyond the final score. An auditable system should make it possible to identify the evidence and assumptions that contributed to uncertainty, understand why a criterion was escalated, and support human review. Such transparency does not eliminate the possibility of bias; it provides a basis for examining whether the routing mechanism, the reference evaluator, or the underlying CV evidence produces unequal outcomes.

**1.3 Aim**

This study is motivated by the two gaps identified above: that escalation in LLM cascades is ordinarily decided for a whole query rather than for its constituent judgments, and that the content of an escalated call has not been treated as a design variable in its own right. It addresses the first directly. Its aim is to evaluate whether criterion-level routing driven by the posterior uncertainty of a hand-specified Bayesian network can reduce reliance on a high-capability reference LLM while retaining agreement with that evaluator and, as a prerequisite, whether that uncertainty signal identifies the decisions on which a local model is in fact unreliable. The corrected evaluation reported here covers one escalation-context policy, a fixed-window slice. Three alternative payload policies are implemented in the accompanying harness, but their re-evaluation under the corrected protocol is left to future work (Section 5.3).

The two research questions addressed here are: 1) Does the posterior uncertainty of the network identify the criterion decisions on which the local 8B model disagrees with the reference evaluator, better than selecting the same number of decisions at random? 2) At the resulting escalation rate, how does routed escalation compare with local-only evaluation and with escalating every criterion, in agreement with the reference evaluator, systematic difference, and token cost?

The design predicted that uncertainty-based routing would substantially reduce reference-model calls while retaining agreement with the reference evaluator, and that its uncertainty signal would concentrate local-model disagreement in the escalated set. The first prediction was supported with respect to calls and tokens; the second was not (Section 4).

Deliberate simplifications constrain the scope. The evaluation uses a single DAG, a fixed five-criterion rubric, and a controlled corpus of synthetic Data Science CVs drawn from a public Kaggle dataset (Section 2.1). The study is therefore a simplified instantiation of the architecture described above, intended to establish whether the mechanism behaves as expected under controlled conditions before extension to larger corpora, multi-DAG structures, or heterogeneous real-world data. The escalation threshold is calibrated on a training subset and locked before evaluation on held-out candidates. The study then compares local-only evaluation, routed escalation under a fixed-window context policy, and escalation of every criterion, using exact agreement, mean absolute difference, Bland–Altman analysis, decision-level agreement, a discrimination analysis of the routing signal, and token expenditure.

*CreateScore* is intended as a cohort-level triage and auditing layer rather than an autonomous hiring system. Its performance relative to a reference LLM does not establish human-level correctness, employment validity, or fairness. Deployment-oriented conclusions therefore require validation using human-labelled and heterogeneous CV data, subgroup-specific performance and fairness analyses, robustness testing under distribution shift, and explicit human oversight.

The remainder of the paper is organized as follows. Section 2 sets out the study design: the provenance and screening of the CV corpus, the five-criterion job order, and the architecture of the Bayesian network and its surrounding pipeline. Section 3 specifies the methodology, including the escalation-context policy evaluated and the evaluation design, and Section 4 reports the resulting agreement, routing-signal and cost measurements. Section 5 relates those results to the questions posed above and to operational screening practice (5.1), sets out the limitations that bound the conclusions (5.2), and identifies the extensions they motivate (5.3). Section 6 introduces a companion planning application that operationalizes the study's measured rates for run estimation, and Section 7 provides the bibliography, the full statistical construction of the network, the reproducibility notes including the defect found in the original evaluation, and usage notes for the application.

## 2. Study Design

The study was designed to isolate a single question: whether criterion-level posterior uncertainty is a workable basis for deciding which screening decisions require a high-capability model. Three design commitments follow from that objective. The corpus is controlled rather than representative, so that variation in agreement can be attributed to the routing mechanism rather than to document heterogeneity. The job order is deliberately small, so that the conditional probability tables remain estimable from a modest training split. And the escalation decision is separated from the escalation payload, so that the two can be varied independently.

### 2.1 About the Dataset

This pilot study evaluation corpus consists of 200 synthetic Data Science curricula vitae obtained from a publicly available Kaggle dataset. The CVs are free-text documents; the pipeline consumes the raw text of each document without access to any structured metadata accompanying it. Real candidate CVs were not available, and hence a synthetic corpus additionally avoids processing personal data belonging to identifiable job applicants.

The synthetic origin of the corpus constrains what the evaluation can establish, and the constraint runs in both directions. Because the documents were generated rather than collected, they exhibit regular formatting and internally coherent career histories; the pipeline is therefore not exercised against the multi-column layouts, inconsistent section headings, scanned-document artifacts, or contradictory self-reporting that characterize real applicant pools. Conversely, that regularity is what permits differences between conditions to be read as consequences of the routing and context-allocation policies rather than of upstream extraction noise. The corpus is best understood as a controlled testbed for the mechanism, not as a sample from any applicant population.

#### 2.1.1 Inclusion-Exclusion Flowchart

A single screening rule was applied: a record was retained if local extraction yielded a valid classification for all five job-order criteria. No further exclusion criteria were imposed, and no records were carried forward, excluded, or re-used from any earlier phase of development; eligibility was assessed once, on the full corpus, in a single pass. All 200 records satisfied the rule, so the eligible cohort is 200 of 200. This complete retention is an observed property of this particular corpus rather than a designed guarantee, and it is a direct consequence of the corpus being synthetic: real CVs frequently contain no evidence whatsoever bearing on a given criterion, a case this corpus does not represent and which is revisited in the Limitations.

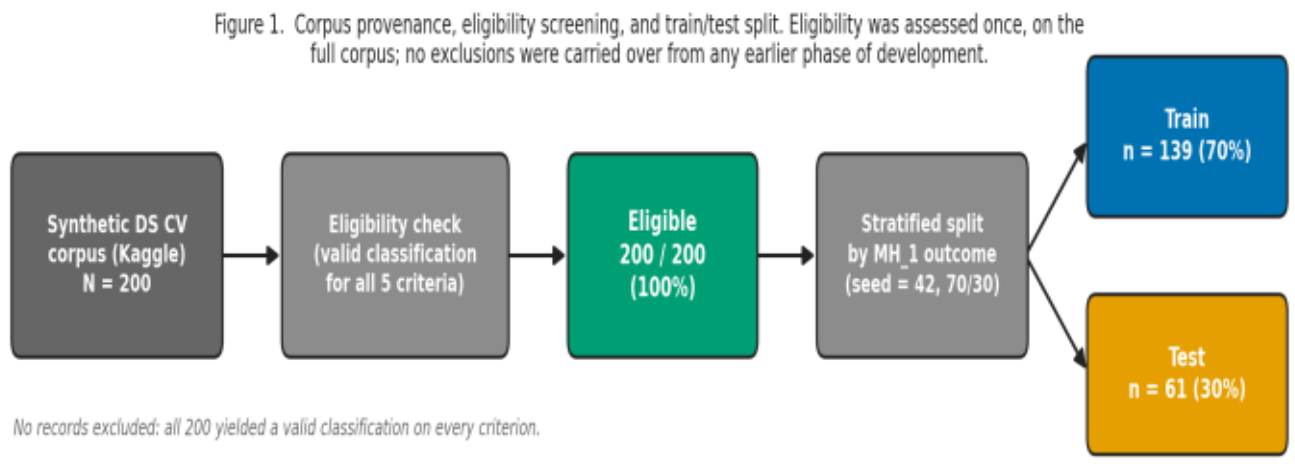

*Figure 1. Corpus provenance, eligibility screening, and the stratified train/test split.*

The eligible cohort was partitioned by a stratified 70/30 split with a fixed random seed (42), yielding 139 training and 61 held-out test candidates (Figure 1). Stratification was performed on the MH_1 outcome because MH_1 operates as a minimum-hire gate: an unstratified split risks producing a training or test fold skewed toward candidates who uniformly clear or uniformly fail that gate, which would distort both the fitted conditional probability tables and the calibrated escalation threshold. The split was computed once and fixed; all subsequent calibration used the training fold alone.

### 2.2 Job Order Payload Design

All candidates were evaluated against one simplified Data Science job order comprising five criteria (Table 1a).

The five-criterion job order was constructed by the author after inspection of the synthetic CV corpus, so that no criterion was satisfied by nearly all or nearly none of the candidates; Table 1b reports the resulting prevalence. Criterion selection therefore reflects the corpus and, because it preceded the train/test split, drew on all 200 documents, although no labels were used in threshold calibration. The experience criterion MH_1 is the exception to the balance: it is deterministic and positive for 82.5% of candidates, and contributes little discriminating information. The split was stratified on MH_1 alone, and the test fold consequently differs from the training fold on the skill criteria (for example, B_4 is positive for 13.1% of test candidates against 29.5% of training candidates). Prompts and payloads were deliberately kept simple under a limited token budget and were not optimized. Results should be read as conditional on this corpus and job order.

Each additional criterion conditioned on the joint latent state multiplies the number of conditional probability table cells to be populated, and a five-criterion rubric keeps every cell supported by a meaningful number of training observations at n = 139. A larger rubric would not invalidate the architecture, but it would shift the binding constraint from the routing question this study addresses to the separate problem of prior elicitation at scale.

| Criterion | Requirement as stated in the job order | Points | Role in the rubric |
|---|---|---|---|
| MH_1 | Minimum 5 years of relevant professional experience | 10 | Minimum-hire gate; also supplies the years-of-experience signal used to derive SS1 |
| B_1 | Proficiency in Python and SQL | 5 | Near-universal baseline skill |
| B_2 | Experience with cloud platforms (AWS, GCP, or Azure) | 5 | Industry-leaning signal; contributes to SS2 |
| B_3 | Research experience (papers, methodology, conference submission) | 5 | Academically-leaning signal; contributes to SS2 |
| B_4 | Advanced ML / statistical modelling, with specific named methods (Bayesian, deep learning, ensembles, PCA, survival or time-series, causal inference, transfer learning); generic mention of "machine learning" alone does not qualify | 5 | Depth-of-method signal; contributes to SS2 |

| Criterion | Points | All 200<br>% positive (inferred / positive) | Training, n = 139<br>% positive (inferred / positive) | Test, n = 61<br>% positive (inferred / positive) |
|---|---|---|---|---|
| MH_1 | 10 | 82.5 (0/165) | 82.7 (0/115) | 82.0 (0/50) |
| B_1 | 5 | 37.0 (9/74) | 41.7 (9/58) | 26.2 (0/16) |
| B_2 | 5 | 39.5 (60/79) | 38.1 (38/53) | 42.6 (22/26) |
| B_3 | 5 | 38.0 (47/76) | 40.3 (36/56) | 32.8 (11/20) |
| B_4 | 5 | 24.5 (22/49) | 29.5 (18/41) | 13.1 (4/8) |

*Table 1(a) (above). The five-criterion job order used throughout the study. Total available score is 30 points; Table 1(b): Prevalence of positive evidence (hit or inferred, which earn equal credit) for each criterion in the full corpus, the training fold and the test fold. The parenthesis gives the number of positives that were classified "inferred" rather than "hit".*

Two features of the payload are load-bearing. First, MH_1 carries double the weight of the behavioral and technical criteria and functions as a gate rather than as an additive component alone, reflecting the practice of treating a stated minimum-experience requirement as qualifying rather than merely contributory. Second, each criterion is scored on a three-way outcome space — *hit*, *inferred*, or *miss* — rather than a binary one. The intermediate category records the common situation in which a CV supplies circumstantial or indirect support for a requirement without stating it explicitly. Preserving that distinction is a precondition for the routing mechanism: a posterior over three outcomes can express genuine ambiguity as elevated entropy, whereas a binary outcome would force such cases into a confident classification and conceal precisely the uncertainty the escalation decision depends upon.

### 2.3 Bayesian DAG Architecture & Pipeline Design

The network contains two latent nodes. SS1 represents a seniority band, derived from the years-of-experience value extracted under MH_1 and discretized as junior (under three years), mid (three to seven years), or senior (over seven years), with a missing value defaulting conservatively to junior. SS2 represents technical background, classified as industry-leaning, academically-leaning, or mixed, and derived deterministically from the candidate's own extracted criterion signals: positive evidence on B_2 without positive evidence on B_3 or B_4 yields an industry-leaning state, the converse yields an academically-leaning state, and all remaining configurations yield a mixed state. Neither derivation invokes a language model and neither does it consult held-out labels, so both are computed identically at training and test time.

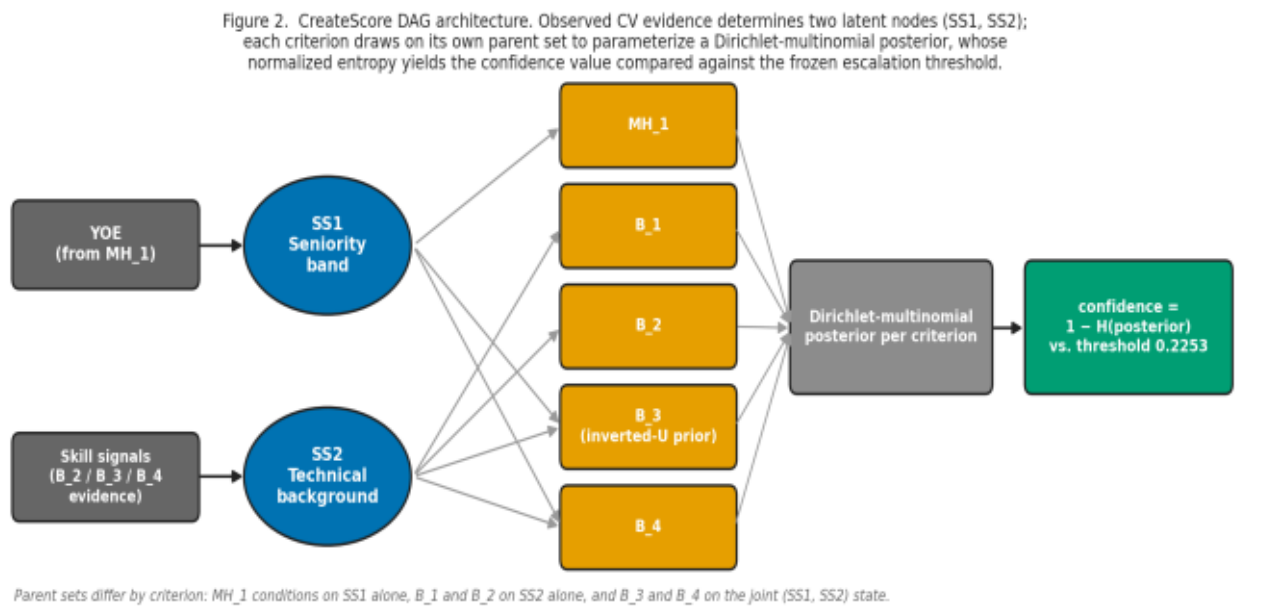

*Figure 2. Network structure, criterion-specific parent sets, and the derivation of the routing signal.*

Each criterion carries a Dirichlet-multinomial conditional probability table over the three outcomes, conditioned on its own parent set rather than on the full latent state (Figure 2). MH_1 conditions on SS1 alone; B_1 and B_2 condition on SS2 alone; B_3 and B_4 condition on the joint (SS1, SS2) state, giving nine cells each. Priors are weakly informative, with hand-elicited pseudo-counts scaled by a global strength parameter fixed at 1.0. The prior shapes encode stated domain assumptions: B_1 is high across all backgrounds, B_2 is moderate and industry-weighted, B_4 increases monotonically in both seniority and academic leaning, and B_3 follows an inverted-U across the joint state, peaking at mid-career academically-leaning candidates and lowest at junior industry-leaning ones, consistent with cumulative-advantage accounts of research output. Posteriors are obtained by adding observed training-fold counts to the prior pseudo-counts.

The routing signal is the normalized entropy of each criterion's posterior mean, expressed as a confidence value of one minus that entropy, so that a value of zero denotes a uniform posterior and a value of one denotes all mass on a single outcome. The escalation threshold was calibrated on the training fold alone by pooling the confidence values of all training criterion-decisions and selecting the quantile that leaves a target proportion of 70% resolved locally. The resulting threshold, 0.2253, was then frozen and applied unchanged to every test-set decision; it was never refit, and no test-set quantity entered its calculation.

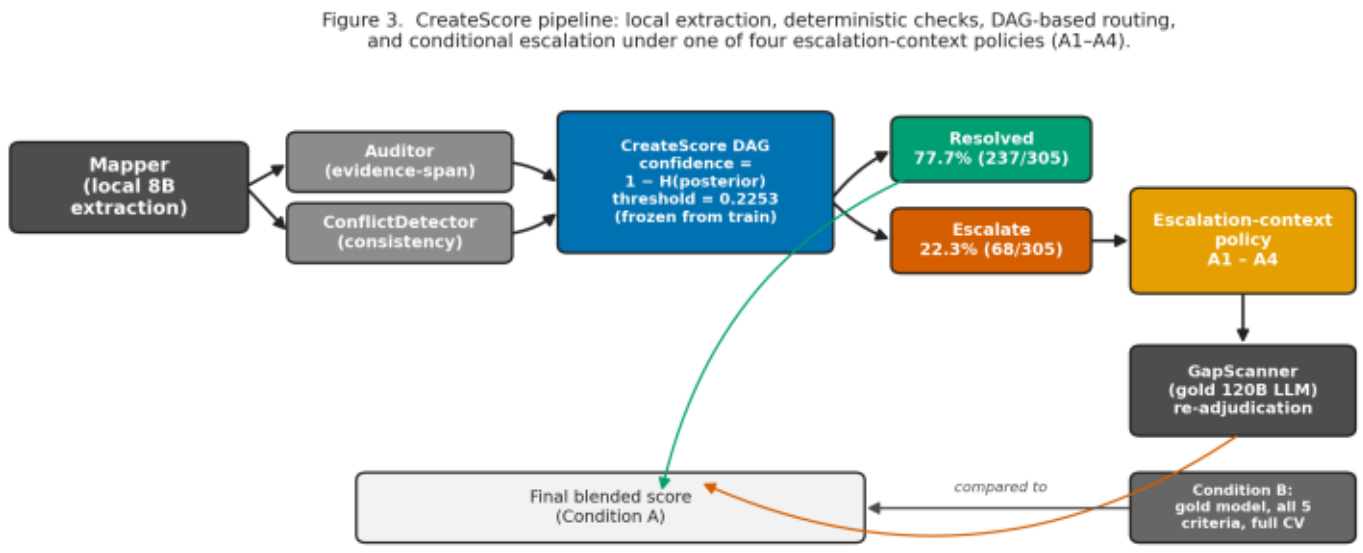

*Figure 3. End-to-end pipeline, showing the routing split and the point at which the escalation-context policy is applied.*

The surrounding pipeline is shown in Figure 3. A Mapper agent performs criterion extraction on each CV using the local 8B model. Two deterministic checks then operate on its output: an Auditor verifies that each claimed classification is supported by a verbatim evidence span (MH_1 is excluded, being a numeric threshold with no corresponding span), and a ConflictDetector tests internal consistency, for example between a stated years-of-experience figure and an implied graduation year. These checks flag extraction artifacts but do not themselves alter the score. The DAG then assigns each criterion-decision to local resolution or to escalation. Escalated decisions are passed to a GapScanner, which re-adjudicates them using the 120B reference model, and the re-adjudicated values are blended with the locally resolved ones to produce the final Condition A score. What context an escalated decision carries to the reference model is determined by the escalation-context policy, which is the variable under test and is specified in Section 3. Condition B, the comparison arm, applies the reference model to all five criteria of every candidate using the CV text truncated at 8,000 characters, and is architecturally identical across all runs.

## 3. Methodology

Criterion extraction was performed by a local 8B model, and escalated decisions were re-adjudicated by a 120B reference model (openai/gpt-oss-120b, served via Groq). The architecture and routing mechanism are as specified in Section 2.3: the network, the local extraction, and the frozen threshold determine the set of escalated decisions (68 of 305), which is therefore identical under every context policy. The evaluation reported in Section 4 uses one context policy, the fixed-window slice (A1; Table 2), in which the escalated criteria of a candidate are sent together with a section-based excerpt of at most 2,500 characters. Three further policies (A2 to A4) are implemented in the accompanying harness. They were run in our preliminary analyses which aren't reproduced here.

| Arm / policy | Context sent to the reference model | Status in this paper |
|---|---|---|
| L: local only | None; no reference call | Evaluated (comparator) |
| A1: fixed window | Escalated criteria only; section-based excerpt of at most 2,500 characters | Evaluated (corrected run) |
| A2: evidence-anchored | Window around the evidence span or a keyword match; fixed-size fallback slices under a 15-decision quota | Implemented; not re-evaluated |
| A3: uncertainty-weighted, uncapped | In each batch of 10, the 40% most uncertain escalated candidates receive the unabridged CV; the rest receive A2 payloads | Implemented; earlier record lost; not re-evaluated |
| A4: uncertainty-weighted, capped | As A3, with the high-context tier capped at 4,000 characters | Implemented; not re-evaluated |
| B: reference | All five criteria; CV text truncated at 8,000 characters | Evaluated (corrected run) |

*Table 2. The comparison arms and escalation-context policies. The 15-decision fallback quota is a shared parameter of the lower-context tier in A2 to A4 and is a heuristic ceiling of roughly 20% of observed escalation volume, not an optimized value.*

### 3.1 Evaluation design

Three arms are compared. Condition L is local-only evaluation: the 8B extraction after the deterministic auditor, with no reference call. Condition A routes escalated criteria to the reference model under the fixed-window policy. Condition B applies the reference model to all five criteria of every candidate, using the CV text truncated at 8,000 characters, and serves as the scoring reference and as the escalate-everything cost baseline. Agreement at candidate level was assessed by exact agreement with Wilson 95% intervals, mean absolute difference (MAD), root-mean-square error (RMSE), and Bland–Altman bias with 95% limits of agreement, with a t-based interval for the bias; the effect of escalation on individual candidates was assessed with an exact two-sided sign test. At decision level, agreement was assessed on the three-way label and on the binary distinction between positive evidence (hit or inferred) and miss, which is the distinction that determines points. The routing signal was evaluated directly: by the area under the receiver operating characteristic curve (AUROC) of one minus confidence for local–reference disagreement, with a candidate-clustered bootstrap interval (5,000 resamples); by the share of disagreements falling in the escalated set against the hypergeometric expectation for a random selection of the same size; and, on escalated decisions, by an exact McNemar test of whether the reference answer corrected or introduced disagreement. Reference calls used a completion limit of 4,000 tokens. A call that was truncated, unparseable, or returned an unrecognised label would have been recorded as a failure and excluded; none occurred. Tokens are those reported by the provider for a single run of each condition, so no run-to-run variance is estimated and token figures are descriptive.

## 4. Analysis & Results

Routing is invariant across conditions. Of 305 criterion-decisions on the test set (61 candidates × 5 criteria), 68 (22.3%) were escalated and 237 (77.7%) were resolved locally. The threshold was calibrated to a 70% target on the training fold, where it resolved 71.7% of decisions under the frozen value of 0.2253. The 6.0-point rise on the test fold appears to reflect its composition. The test fold contains fewer candidates in the academically-leaning latent state than the training fold (21% against 37%), and resolution rose for B_1 (79% against 63%), B_3 (95% against 82%) and B_4 (66% against 56%), while falling for B_2 (49% against 57%). We did not test this explanation directly. At candidate level, 44 of 61 candidates (72.1%) required at least one reference-model call, and 17 (27.9%) were resolved entirely locally. The gap between the criterion-level and candidate-level savings is a direct consequence of decomposed routing: a single uncertain criterion is sufficient to draw a candidate into the expensive tier, so savings dilute as they aggregate upward.

| Measure (reference = Condition B, n = 61) | Local only (8B + auditor) | CreateScore, routed (A1) |
|---|---|---|
| Exact agreement, n (%) | 20 (32.8%) | 26 (42.6%) |
| Wilson 95% CI | 22.3% to 45.3% | 31.0% to 55.1% |
| Mean absolute difference | 4.43 | 3.93 |
| Root-mean-square error | 5.80 | 5.65 |
| Bias, mean (score − reference), 95% CI | +1.80 (0.38 to 3.23) | +0.66 (−0.79 to 2.11) |
| Bland–Altman 95% limits of agreement | −9.00 to +12.60 | −10.35 to +11.66 |
| Candidates closer to the reference than local only | n/a | 9 closer, 3 further, 49 tied (sign test p = 0.146) |

*Table 3. Candidate-level agreement with the complete reference condition (n = 61). Bias is the mean paired difference; positive values denote over-crediting relative to the reference. Confidence intervals for bias are t-based (60 degrees of freedom), using the sample standard deviation of the paired differences. The sign test is exact and two-sided, on the 12 candidates whose absolute difference changed.*

### 4.1 Results

**Candidate-level agreement**. Against the complete reference, local-only evaluation agreed exactly for 20 of 61 candidates (32.8%; Wilson 95% CI 22.3–45.3%), with MAD 4.43, RMSE 5.80 and a bias of +1.80 points (95% CI 0.38 to 3.23), indicating that the local model over-credits relative to the reference (Table 3). Routed escalation raised exact agreement to 26 of 61 (42.6%; 31.0–55.1%), lowered MAD to 3.93 and RMSE to 5.65, and reduced the bias to +0.66 (−0.79 to 2.11), an interval that includes zero. Escalation moved a candidate's score closer to the reference for 9 candidates, further from it for 3, and left 49 unchanged (exact sign test p = 0.146). The 95% limits of agreement remained wide, from −10.35 to +11.66 points on the 30-point scale.

**Token expenditure**. Table 3b reports provider-recorded tokens. Adjudicating every criterion with the reference model consumed 121,174 tokens (93,220 input and 27,954 output, of which 23,584 were reasoning tokens), or 1,986 per candidate. Routed escalation under the fixed-window policy consumed 42,170 tokens (35,896 input and 6,274 output; 4,820 reasoning), or 691 per candidate, a reduction of 65.2% (61.5% of input and 77.6% of output tokens). The output reduction is larger because escalated calls request only the escalated criteria and so incur less reasoning. These are single-run totals without replication and are descriptive.

**Decision level**. The local model and the reference disagreed on whether a criterion was satisfied (positive evidence against a miss) for 57 of 305 decisions (18.7%; 14.7–23.4%) and on the three-way label for 70 (23.0%; 18.6–28.0%). The disagreement was not concentrated in the escalated set: it was 11 of 68 (16.2%; 9.3–26.7%) among escalated decisions and 46 of 237 (19.4%; 14.9–24.9%) among decisions resolved locally (Table 3a). Eleven of the 57 binary disagreements (19.3%) fell in the escalated set, against 12.7 expected for a random selection of 68 decisions (hypergeometric P(X ≥ 11) = 0.78; 95% range under random selection, 7 to 18). The area under the ROC curve for one minus confidence as a predictor of local–reference disagreement was 0.473 (candidate-clustered bootstrap 95% CI 0.392–0.557). In this sample the uncertainty signal carried no detectable information about local-model error. Among the 68 escalated decisions the local model agreed with the reference on 57 (83.8%) and the escalated answer on 65 (95.6%): escalation corrected 11 disagreements and introduced 3 (exact McNemar p = 0.057). Four-fifths of the disagreements (46 of 57) remained among locally resolved decisions.

**Reference integrity**. All 61 Condition B calls and all 44 Condition A calls in the corrected evaluation terminated normally (finish reason "stop") and returned a valid classification for every requested criterion; none was replaced by a local label. Reasoning tokens accounted for 84.4% of Condition B completion tokens (23,584 of 27,954) and 76.8% of Condition A completion tokens (4,820 of 6,274). The mean of 387 reasoning tokens per full-rubric reference call leaves almost nothing of a 400-token completion limit for the visible answer, which is the mechanism of the defect described in Section 7.2 (Table 3c).

| Local-model disagreement with the reference | Escalated (n = 68) | Resolved locally (n = 237) | All (n = 305) |
|---|---|---|---|
| Binary (positive evidence vs miss) | 11 (16.2%)<br>9.3% to 26.7% | 46 (19.4%)<br>14.9% to 24.9% | 57 (18.7%)<br>14.7% to 23.4% |
| Three-way label | 15 (22.1%)<br>13.8% to 33.3% | 55 (23.2%)<br>18.3% to 29.0% | 70 (23.0%)<br>18.6% to 28.0% |
| Router discrimination (AUROC of 1 − confidence) | 0.473 (95% CI 0.392 to 0.557) | | |
| Disagreements captured by escalation | 11 of 57 (19.3%); 12.7 expected at random; 95% random range 7 to 18; P(X ≥ 11) = 0.78 | | |
| Effect of the escalated call on escalated decisions (binary) | Local agrees with reference 57 of 68 (83.8%); after escalation 65 of 68 (95.6%); corrected 11, introduced 3; McNemar p = 0.057 | | |

| Condition | Calls | Input tokens | Output tokens (reasoning) | Total | Per candidate |
|---|---|---|---|---|---|
| B: reference on all criteria | 61 | 93,220 | 27,954 (23,584) | 121,174 | 1,986 |
| A1: routed escalation | 44 | 35,896 | 6,274 (4,820) | 42,170 | 691 |
| Reduction | 27.9% of calls (77.7% of criterion decisions) | 61.5% | 77.6% | 65.2% | 65.2% |

| Parameter | Original harness | Corrected harness |
|---|---|---|
| Completion-token limit | 400 | 4,000 |
| Truncated or unparseable response | Replaced by the 8B labels; zero tokens recorded; nothing logged | Recorded as a failure; never replaced |
| Unrecognised label | Coerced to "miss" | Recorded as a failure |
| Diagnostic replay of the Condition B call (first 12 test candidates) | 7 of 12 ended with finish reason "length" and could not be parsed; reasoning tokens 339 to 398 against the limit of 400 | n/a |
| Full corrected run | n/a | 0 of 61 Condition B and 0 of 44 Condition A calls failed; all finished with reason "stop" |
| Condition B input tokens recorded, as % of the corrected run's 93,220 | 39.6%, 31.9%, 28.8% (three earlier runs) | 100% by construction |
| Consequence | Condition B was largely composed of local labels; agreement and cost figures invalid | Reference condition complete |

*Table 3(a-c) Decision-level agreement and routing-signal discrimination. Intervals are Wilson 95% for proportions. The AUROC interval is a candidate-clustered bootstrap with 5,000 resamples; Provider-recorded token usage in the corrected evaluation. One call per candidate in each condition; Condition A calls were made for the 44 candidates with at least one escalated criterion; Reference-integrity audit. Input tokens depend only on the prompt and CV text, which were identical across runs, so the shortfall in the earlier runs implies that most Condition B calls returned no usable response.*

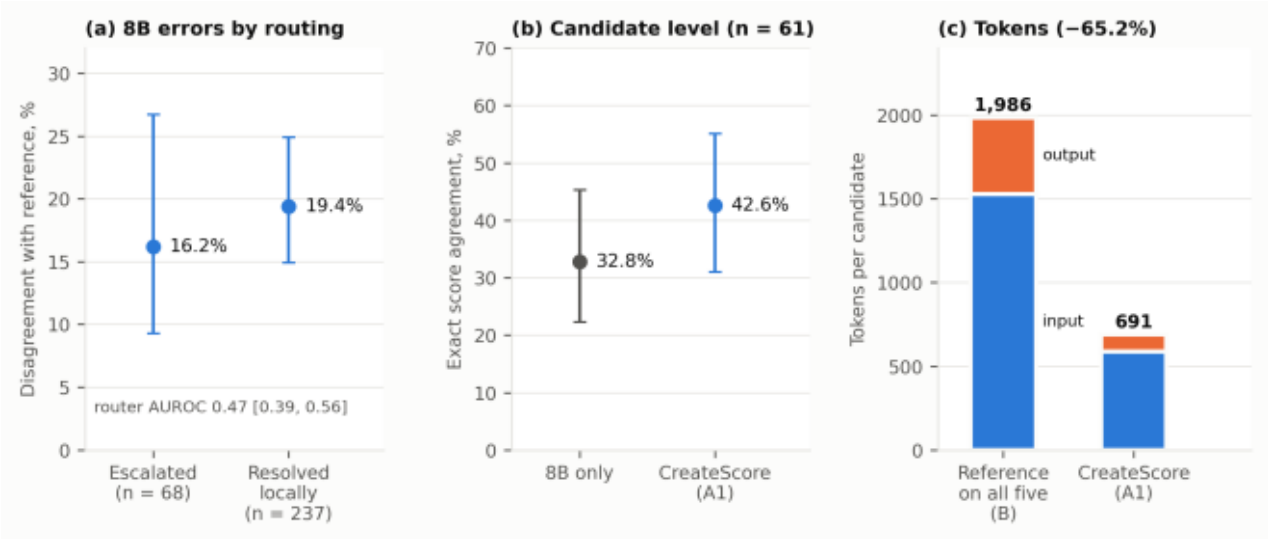

*Figure 4. Results on the 61 test candidates. (a) Disagreement between the 8B model and the reference on escalated and locally resolved decisions (binary: positive evidence against miss), with Wilson 95% intervals; the AUROC interval is a candidate-clustered bootstrap. (b) Exact score agreement with the reference for local-only and routed evaluation, with Wilson 95% intervals. (c) Provider-recorded tokens per candidate, single run per condition.*

### 4.2 Interpretation.

Three results stand, and one does not. First, the mechanism does what a cascade is designed to do with respect to cost: 77.7% of criterion decisions and 65.2% of tokens were removed relative to adjudicating every criterion with the reference model. Second, escalation moved the evaluation toward the reference. Exact agreement rose from 32.8% to 42.6%, and the bias fell from +1.80 points, whose interval excludes zero, to +0.66, whose interval does not; but at n = 61 the intervals for agreement overlap and the paired shift is not significant ($p = 0.146$), so the improvement is suggestive rather than established. Third, the local model's over-crediting is the dominant source of candidate-level error, and it is the escalated reference answers that corrected it: on the 68 escalated decisions the reference answer corrected 11 disagreements and introduced 3.

What does not stand is the claim that motivated the architecture. The posterior uncertainty did not identify the decisions on which the 8B model erred. Disagreement was no more frequent among escalated decisions (16.2%) than among those resolved locally (19.4%); the area under the ROC curve was 0.47, with an interval that straddles 0.5; and the share of disagreements captured by escalation (11 of 57, 19.3%) is close to the 12.7 expected if the same number of decisions had been chosen at random. The gain in agreement is therefore attributable to the reference call, not to the targeting, and four-fifths of the local–reference disagreements (46 of 57) remained among decisions the network judged confident. A plausible but untested explanation lies in the construction of the signal. Confidence is the normalized entropy of a posterior mean within a (criterion, stratum) cell and takes only a small number of distinct values, so it describes how predictable a criterion is within a stratum rather than whether this candidate's extraction is reliable. The evidence here is compatible with that account but does not establish it.

## 5. Discussions

### 5.1 Inference

The results support a narrower claim than the one this study set out to test. Routing criterion-level decisions on the posterior uncertainty of a hand-specified Bayesian network removed 77.7% of reference-model calls and 65.2% of reference-model tokens, and the escalated answers moved candidate scores toward the reference. The inspectable posterior makes each routing decision traceable, which has value for a screening operation that must explain why a judgment was, or was not, sent for further review. Traceability, however, is not discrimination. In this sample the posterior did not direct the expensive resource toward the judgments on which the local model was least reliable, and a reader should not read a high-confidence local label as evidence that the underlying extraction is trustworthy.

The more transferable finding is methodological. A cascade harness whose escalation path substitutes the cheap model's answer when the expensive call fails, without recording the failure, will report agreement with a reference that is in part itself. The substitution inflates agreement, deflates the apparent cost of the reference, and varies from run to run in a way that is easily mistaken for provider non-determinism. We encountered all three effects: our preliminary analyses attributed a variation of about 35% in reference-model token usage to provider-side non-determinism, when it reflected truncation. Reasoning models compound the hazard because reasoning tokens count against the completion limit, and a limit adequate for a conventional model (400 tokens) was exhausted by the reference model's reasoning alone (a mean of 387 tokens per full-rubric call). We recommend that cascade evaluations log finish reasons and failure counts for every arm, treat failed calls as missing and never as substitutable, and check that reference-condition token totals scale with the number of calls and the lengths of their payloads.

Payload design, the second gap that motivated this study, remains open. The earlier four-policy comparison cannot be relied upon, and we draw no conclusion about the relative merit of fixed-window, evidence-anchored and uncertainty-weighted payloads. As an exploratory observation only, the scoped escalation calls of that evaluation were largely unaffected by the defect (the fixed-window run's recorded input and output tokens were 94.5% and 91.0% of the corrected re-run's), and the candidate scores produced under policies A1, A2 and A4 coincided for 58 or 59 of 61 candidates. If that pattern survives a corrected comparison, payload would matter little at this scale; the present data do not establish it.

### 5.2 Limitations

The corpus is 200 synthetic CVs evaluated against one simplified five-criterion rubric and a single reference model. These results characterize a controlled setting, not a validated production system, and three limitations follow directly.

**Missingness:** every candidate here had extractable evidence for all five criteria, whereas real CVs routinely lack evidence for a criterion entirely — a genuinely absent signal rather than a harder-to-find one. A deployment would need an explicit not-evaluable state distinct from a miss, and a policy for it, rather than allowing a low-information case to present as a confident negative.

**Idealization:** the graph was elicited for one job family and should not be assumed to transfer to a different rubric without re-elicitation, and the synthetic documents do not exercise the noisy upstream extraction — inconsistent formatting, multi-column layouts, contradictory self-reporting — that precedes the network in any real pipeline.

**Attribution**: the evaluation tests whether the uncertainty signal discriminates local-model error (AUROC, capture against the hypergeometric expectation) and finds no evidence that it does. It does not include an arm in which a random 22.3% of decisions is escalated and re-adjudicated, nor a comparison with simpler uncertainty signals such as agreement across repeated local samples or token probabilities, and the capture analysis is limited to the 305 decisions of one test fold.

**Statistical assumptions.** The Dirichlet-multinomial tables assume exchangeability of candidates within each latent stratum, which is reasonable under synthetic generation but not guaranteed for an applicant pool with channel-specific batch effects. The frozen threshold assumes train and test are drawn from the same distribution; a drifting pool would require periodic recalibration. The paired metrics treat the 0–30 scale as additive across five unequally weighted criteria, which is defensible but non-trivial for what is at heart a discrete rubric. The reference is a single draw of a 120B language model, not human adjudication, so agreement measures consistency with a larger model and not correctness; its run-to-run reproducibility under the corrected harness was not measured. The corrected evaluation consists of one run of each condition, one split and 61 test candidates, so all intervals are wide.

Finally, the projection that savings grow with corpus size assumes a fixed threshold; the procedure used here re-targets a 70% resolution rate, which would absorb sharper posteriors rather than convert them into additional savings.

### 5.3 Future Research Scope

The clearest remaining lever follows from the main negative result: (a) Candidate-specific uncertainty. The posterior confidence used here varies mainly between criteria and strata, and it did not locate the local model's errors (Section 4). Signals that vary within a stratum, such as agreement across repeated local samples or local token probabilities, should be tested against the posterior, with risk–coverage curves as the common yardstick. (b) Re-evaluation under the corrected harness. The three payload policies not re-run here (A2 to A4) should be re-evaluated with replicates, a repeated reference condition to measure its reproducibility, and a matched-rate random-routing arm. (c) Structured uncertainty payloads. All four policies vary only the volume and selection of CV text; none transmits what the network already knows — its posterior confidence for the criterion in question and the partial signal behind it. Pairing a narrow text budget with an explicit statement of that posterior is untested and is a direct route to closing the gap between call and token savings. (d) Active updating. Every escalation produces a reference-model label that is currently discarded; feeding it back into the posteriors would let the system escalate less as it accumulates experience, turning a static cost profile into an improving one. Together these directions turn today's result into a trajectory rather than a fixed point.

Five further directions are more incremental. (e) Cross-candidate batching of escalated criteria into single reference-model calls would amortize per-call overhead at scale. (f) Threshold characterization: the 70% target is one point on an accuracy–efficiency frontier, and mapping where along that frontier agreement begins to degrade would give deployers a principled dial rather than a fixed setting. (g) Richer latent structure, or a set of smaller connected graphs specialized by job family, would localize uncertainty more precisely than one global structure and should permit narrower payloads, at the cost of more elicitation and more data. (h) Stakes-weighted escalation — deferring high-point criteria more readily than low-point ones — offers a route to narrowing the wide per-candidate limits of agreement reported in Table 3, which remain an honest boundary of the current design. (i) Human adjudication of a stratified subset of criterion decisions would separate disagreement with a larger model from error. Beyond these, validation on real heterogeneous CVs with human-adjudicated labels, a second independent reference model, and subgroup fairness analysis remain prerequisites for any deployment claim.

## 6. Streamlit Application

A companion Streamlit application accompanies this paper, operationalizing the study's measured rates for run planning (GitHub). Given a candidate count — entered directly, or inferred from an uploaded job-order payload — it computes the stratified train/test split, lays out the resulting run schedule over the test set, projects wall-clock run time, and scales the observed escalation rates and per-candidate token costs to the new pool size to produce an escalated-criterion count and a coarse reference-token comparison against an escalate-everything baseline.

Its purpose is practical rather than confirmatory. The application performs no scoring, routing, or posterior computation; it does not reproduce the network, and it makes no claim to predict agreement. Every constant it uses — the 70% train fraction, the chunk size of 10, the per-candidate timing (about 16 seconds when unthrottled and 34 seconds including provider rate-limit pauses), the 22.3% criterion-level and 72.1% candidate-level escalation rates, and the reference-model token costs per candidate (1,986 for full-rubric adjudication and 958 for a routed call) — is a value measured in the corrected evaluation and exposed as an adjustable input rather than a fixed assumption, so that a practitioner planning a differently sized or differently resourced run can re-calibrate it rather than inherit this study's conditions. Its limitations are those of its inputs: the timing and token constants derive from a single observed run, the escalation rates depend on the corpus and the calibrated threshold (the training fold resolved 71.7% of decisions locally against 77.7% on the test fold), and the cost projection assumes escalation behaves proportionally at larger scale, which the scaling argument in Section 5.2 explicitly qualifies. Usage details are given in Section 7.3.

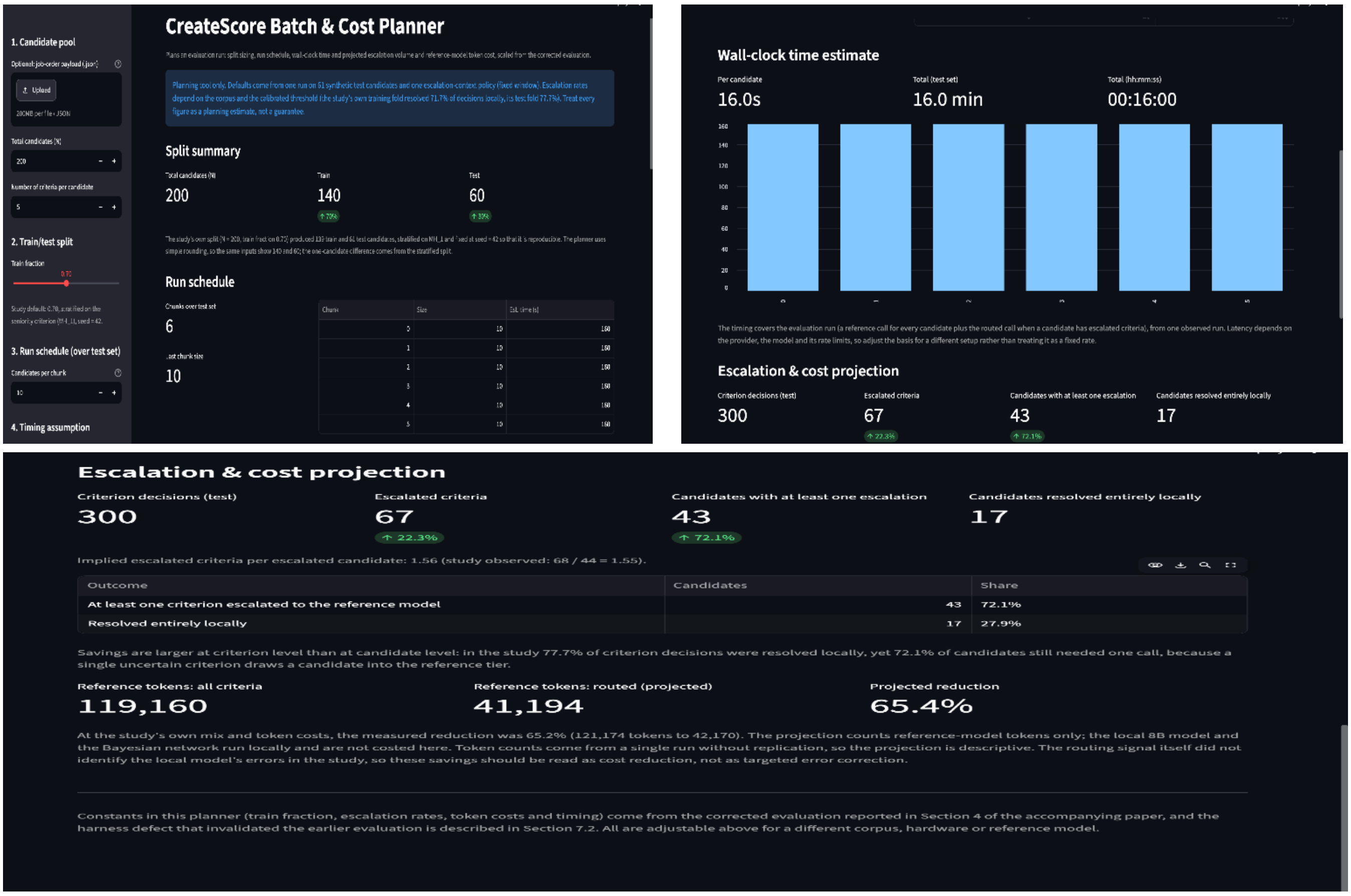

*Figure 5: CreateScore Interactive App interface for planning and study design.*

## 7. Appendix

### 7.2 Statistical construction of the DAG

**Latent variables.** SS1 is derived from the years-of-experience value extracted under MH_1 and banded as junior (< 3 years), mid (3–7), or senior (> 7), with a missing value defaulting to junior. SS2 is derived deterministically from the candidate's own extracted signals: positive evidence (hit or inferred) on B_2 without positive evidence on B_3 or B_4 gives an industry-leaning state, the converse gives an academically-leaning state, and every other configuration gives a mixed state. Neither derivation calls a language model or consults held-out labels.

| Criterion | Parent set | Cells | Prior shape on the hit outcome |
|---|---|---|---|
| MH_1 | SS1 (seniority) | 3 | Monotone increasing in seniority |
| B_1 | SS2 (background) | 3 | High across all backgrounds (near-universal skill) |
| B_2 | SS2 (background) | 3 | Moderate to low; highest for industry leaning candidates |
| B_3 | SS1 × SS2 (joint) | 9 | Inverted-U: peak at (mid, academic), lowest at (junior, industry) |
| B_4 | SS1 × SS2 (joint) | 9 | Monotone increasing in both seniority and academic leaning |

*Table 4. Parent sets and prior shapes. Full pseudocount values are in build_priors() in the accompanying source.*

**Priors, posteriors, and the routing signal.** Each cell holds a Dirichlet prior over the three outcomes (hit, inferred, miss) with hand-elicited pseudocounts scaled by a global strength parameter fixed at 1.0, giving weakly informative cell totals of roughly two to four. The posterior concentration is the elementwise sum of the prior pseudocounts and the observed training-fold counts, and the posterior mean is that vector normalized to sum to one. Confidence is one minus the entropy of the posterior mean normalized by log 3, so that zero denotes a uniform posterior and one denotes all mass on a single outcome. To calibrate the escalation threshold, the confidence values of all training criterion-decisions were pooled and sorted, and the threshold set at the quantile leaving a target proportion of 0.70 resolved locally. The resulting value, 0.2253, was frozen and applied unchanged to the test fold. A decision resolves locally when confidence is at or above the threshold and escalates otherwise.

**Reproducibility.** The corrected evaluation uses the harness described in Section 3, with a completion limit of 4,000 tokens, explicit failure recording, and an on-disk cache of every call, its finish reason and its token usage. Condition B and the fixed-window policy were run once, in a single session, on the 61 test candidates, and the escalated set was verified to match the frozen routing for every candidate. The corrected run is reproduced by rerun_fixed.py followed by analyze_rerun.py;

```
A1  --mode live --gold gpt-oss-120b --escalation-mode full-cv
A2  --mode live --gold gpt-oss-120b --escalation-mode evidence --max-fallback 15
A3  --mode live --gold gpt-oss-120b --escalation-mode batch-hybrid --batch-size 10 --batch-full-cv-frac 0.40
A4  --mode live --gold gpt-oss-120b --escalation-mode batch-hybrid --batch-size 10 --batch-full-cv-frac 0.40 \
        --batch-full-cv-chars 4000
```

*Defaults throughout: RANDOM_SEED = 42, TRAIN_FRAC = 0.70, TARGET_RESOLVE_RATE = 0.70, EVIDENCE_CONTEXT_LINES = 3, EVIDENCE_FALLBACK_CHARS = 2500, EVIDENCE_DEGRADED_CHARS = 800.*

Defect in the original evaluation. The first evaluation of this system (policies A1 to A4) used a harness with a completion limit of 400 tokens in which a response that was truncated, unparseable or invalid was replaced by the local model's labels, with zero tokens recorded and no entry logged; an unrecognised label was coerced to "miss". Because the reference model reasons before answering, the limit was frequently exhausted. In a diagnostic replay of the Condition B call for the first twelve test candidates, seven ended with finish reason "length" and could not be parsed, with reasoning tokens of 339 to 398 against the limit of 400. The input-token totals recorded for Condition B in the three retained runs (36,899, 29,691 and 26,883) were 39.6%, 31.9% and 28.8% of the 93,220 recorded when every call completed, which implies that most Condition B calls in those runs returned no usable response. Agreement, bias and token figures from that evaluation are therefore invalid and are not reported; the variation in reference-model token usage that our preliminary analysis attributed to provider non-determinism was an artefact of the substitution.

### 7.3 Usage of the Streamlit app

The application is a single Python file with two dependencies, launched from the directory containing it:

```
pip install streamlit pandas
streamlit run createscore_planner_app.py
```

It serves the interface at a local address, by default port 8501. Inputs are entered in the sidebar in five groups: the candidate pool, either as a count or as an uploaded job-order JSON from which the candidate and criteria counts are inferred; the train fraction; the batch size; the per-candidate timing constant; and the criterion- and candidate-level escalation rates. The main panel returns the resulting split, a per-batch schedule with an estimated duration for each batch, a total wall-clock projection, a projected count of escalated criteria, a tier breakdown scaled to the new pool, and a token-cost comparison against an escalate-everything baseline.

The default timing of about 16 seconds per candidate is the mean interval between consecutive candidates in the unthrottled stretches of the corrected run; across the whole run, including two provider rate-limit pauses of about 19 minutes in total, it averaged 34 seconds per candidate (2,081 seconds across 61 candidates). Both figures are single observations and will vary with reference-model latency, network conditions, provider limits and the escalation mix of the corpus at hand; the time basis can be switched or replaced rather than inherited. The token projection multiplies the study's measured reference-model tokens per candidate by the projected number of candidates and counts reference-model tokens only. It is a single-run, descriptive planning estimate and not a cost guarantee.

**Data and code availability**. The 200 synthetic CVs are from the publicly available Kaggle dataset cited in Section 2.1 and are not redistributed here. The Streamlit planner is available at< GitHub link>.